\documentclass[runningheads]{llncs}

\usepackage{cite}

\usepackage[utf8]{inputenc}
\usepackage[english]{babel}
\usepackage[T1]{fontenc}
\usepackage{float}
\usepackage{svg}
\usepackage{tikz}
\usepackage{hyperref}

\hypersetup{
  colorlinks   = true, 
  urlcolor     = blue, 
  linkcolor    = red, 
  citecolor    = green 
}

\newcommand\copyrighttext{%
  \scriptsize This version of the article has been accepted for publication, after peer review and is subject to Springer Nature’s AM terms of use, but is not the Version of Record and does not reflect post-acceptance improvements, or any corrections. The Version of Record is available online at: \url{http://dx.doi.org/10.1007/978-3-030-73959-1_19}. © Springer Nature Switzerland AG 2021
F. Heintz et al. (Eds.): TAILOR 2020, LNAI 12641, pp. 212–219, 2021.}
\newcommand\copyrightnotice{%
\begin{tikzpicture}[remember picture,overlay]
\node[anchor=south,yshift=50pt] at (current page.south) {\fbox{\parbox{\dimexpr\textwidth-\fboxsep-\fboxrule\relax}{\copyrighttext}}};
\end{tikzpicture}%
}

\begin{document}

\title{Process-To-Text: a framework for the quantitative description of processes in natural language}
\titlerunning{Quantitative description of processes in {N}atural {L}anguage}

\author{Yago Fontenla-Seco\inst{1}\orcidID{0000-0002-7566-0164} \and
 Manuel Lama\inst{1}\orcidID{0000-0001-7195-6155}
 \and
 Alberto Bugarín\inst{1}\orcidID{0000-0003-3574-3843}
}

\authorrunning{Y. Fontenla-Seco et al.}

\institute{Centro Singular de Investigación en Tecnoloxías Intelixentes (CiTIUS)\\
 Universidade de Santiago de Compostela, Spain. \\
 \email{\{yago.fontenla.seco, manuel.lama, alberto.bugarin.diz,\}@usc.es}
}

\maketitle

\begin{abstract}
 
 In this paper we present the Process-To-Text (P2T) framework for the automatic generation of textual descriptive explanations of processes. P2T integrates three AI paradigms: process mining for extracting temporal and  structural information from a process, fuzzy linguistic protoforms for modelling uncertain terms, and natural language generation for building the explanations. A real use-case in the cardiology domain is presented, showing the potential of P2T for providing natural language explanations addressed to specialists.
 
 \keywords{Process mining \and
  Natural Language Generation \and
  Explainable AI
 }
\end{abstract}

\vspace*{-0.75cm}

\copyrightnotice

\section{Introduction}
\label{sec:introduction}
%
Processes constitute a useful way of representing and structuring the activities in information systems. The Process Mining field offers techniques to discover, monitor and enhance real processes extracted from the event logs generated by processes execution, allowing to understand what is really happening in a process, which may be different from what designers thought  \cite{Aalst16}.
Process models are usually represented with different notations that represent in a graphical manner the activities that take place in a process as well as the dependencies among them.
Process models tend to be enhanced with properties such as temporal information, process execution-related statistics, trends of process key indicators, interactions between the resources involved in the process execution, etc.
Information about these properties is shown to users through visual analytic techniques that help to understand what is happening in the process execution from different perspectives.
However, in real scenarios process models are highly complex, with many relations between activities, which make them nearly impossible to be interpreted and understood by the users \cite{Aalst16}.
Furthermore, the amount of information that can be added to enhance the process description is also very high, and it is quite difficult for users to establish its relation with the process model.
Additional analytics which summarize quantitative relevant information about a process, such as frequent or infrequent patterns, that make it easier to focus on finding usual or unexpected behaviors, are also very useful. \cite{ChapelaCampa2019}

Nevertheless, their correct interpretation is usually difficult for users, since they need to have a deep knowledge about process modelling and process related visual analytics.
In this regard, some approaches have been described to automatically generate natural language descriptions of a process aiming to provide users with a better understanding of it \cite{Leopold2012}.
These descriptions aim to explain in a comprehensible way, adapted to the user information needs, the most relevant information of the process.
In general, textual information is complementary to process model visualization, which is the usual way of conveying information to users.

Natural Language Generation (NLG) \cite{Reiter2000, Gatt2018} provides different methods for generating insights on data through natural language.
Its aim is to provide users with textual descriptions that summarize the most relevant information of the data that is being described.
Natural language is an effective way of conveying information to humans because i) it does not rely on human capability to identify or understand visual patterns or trends; and ii) it may include uncertain terms of expressions, which are very effective for communication \cite{aramossoto2016prai}.
Research suggests that in some specialized domains knowledge and expertise are required to understand graphical information \cite{petre95} and proves that specialists can take better decisions based on textual summaries than on graphical displays \cite{Law2005}.
With the integration of NLG and process model and analytics visualization, users with limited expertise on process modeling and analysis are provided with an AutoAI tool that allows them to understand the relevant information about what is really happening within a process.

In the state-of-the-art, techniques for process textual description focus on two perspectives.
On the one hand, the Control-Flow perspective aims to align a Business Process Model Notation (BPMN) representation of a process model with its corresponding textual description \cite{Leopold2012,Leopold2014,Aa2015,SanchezFerreres2017}.
The aim of this proposal is to, first, facilitate users to understand the process model structure and the dependencies between activities and how resources are used and second, detect inconsistencies in the process model such as missing activities, activities in different order, or misalignments in order to maintain a stable process model through the different stakeholders of an organization.
The main drawback of these approaches is that they focus on hand-made processes (not discovered from real data) with a well-defined structure, and draw all their attention in the validation step of the process design phase.
Preventing its application to process models extracted from real-life data, usually unstructured, with many relations between activities, including frequent loops, parallels and choices.
On the other hand, Case Description techniques focus on generating textual descriptions about the execution of single activities or activity sequences that have been registered in an event log \cite{Dijkman2017}.
The underlying process model is not discovered from the event log, therefore neither the process structure is considered nor the relations between activities. So, these last techniques do not provide a faithful description of what has happened during the process execution.

In this paper, we present the Process-To-Text (P2T) framework, a process mining-based framework (the real process model is discovered from the event data) for the automatic generation of natural language descriptions of processes. Descriptions include information from both the control-flow, case and specially time perspectives, the later being usually neglected in the literature.
P2T is based on a Data-to-Text (D2T) architecture \cite{Reiter2007} using linguistic protoforms (as a way to handle imprecision) that will be generated into natural language texts following a hybrid template-based realization approach.

\section{P2T: a framework for textual description of processes}
\label{sec:framework}

\begin{figure}[!t]
 \centering
 \includegraphics[width=0.75\linewidth]{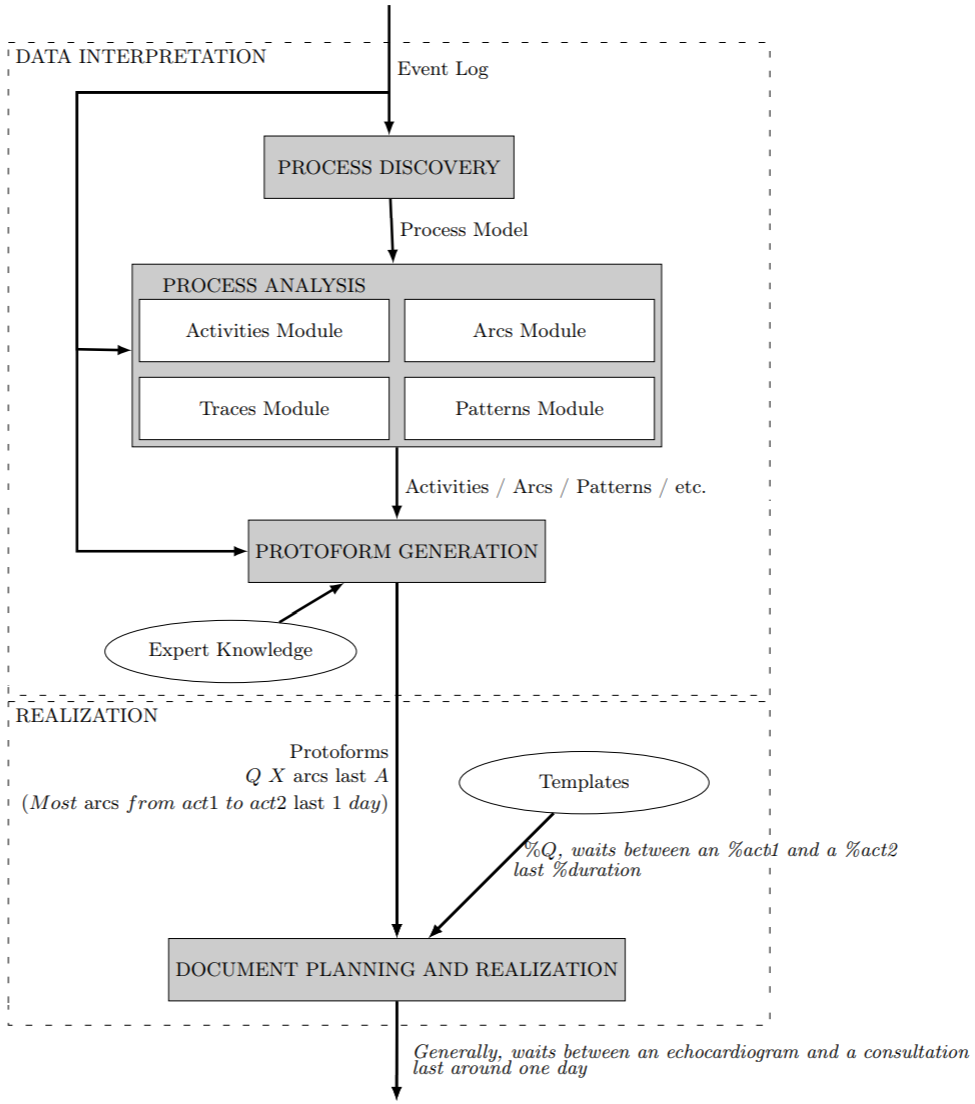}
 \caption{Framework for the linguistic description of processes}
 \label{fig:architecture}
\end{figure}

Fig. \ref{fig:architecture} depicts the Process-To-Text (P2T) framework, for the automatic generation of textual descriptive explanations of processes.
This framework is based on the most widely used architecture for D2T systems \cite{Reiter2007}, which defines a pipeline composed of four stages: \textit{signal analysis}, \textit{data interpretation}, \textit{document planning}, and \textit{microplanning and realization.}
P2T does not include the \textit{signal analysis} stage since data input are not numerical, but event logs.
Also, the \textit{document planning} stage is subsumed in the \textit{microplanning and realization} stage.

\subsubsection{Process discovery.} The input of the data interpretation stage is an \textit{event log}, defined as a multiset of traces. A trace is a particular execution of a process (i.e., a case), and it is represented as an ordered sequence of events, being an event the execution of an activity, which contains context information about the execution said activity (e.g. timestamp, the case it belongs to, the resources involved in its execution, the process variables modified, etc.).
However, the event log itself is not used as a direct input to generate the descriptions.
Firstly, the process model has to be discovered from the event log by applying a process mining algorithm \cite{Aalst16, 8368306}, which, traditionally have followed heuristic \cite{heuristics-miner}, inductive \cite{10.1007/978-3-642-38697-8_17}, or evolutionary computation \cite{DBLP:journals/isci/Vazquez-BarreirosML15} based approaches.

\subsubsection{Process Analysis.}  Once the process is mined, the log is replayed (each trace is played over the discovered model) \cite{Aalst16}.
This gives us both temporal and frequency-based information about activities, arcs (relations between activities) and traces that can be as well used to extract frequent and infrequent behavioral patterns \cite{ChapelaCampa2019}.
Then, this information is summarized into indicators (e.g. average duration and frequency of the relation between activities, average and mode duration of a path, changes of mean duration of an activity within a period, etc.) which are computed in the modules depicted in Figure \ref{fig:architecture} in the process analysis phase.
This phase is part of the framework and indicators are computed for any case or domain.

\subsubsection{Protoform generation.} A protoform \cite{Zadeh2002} is an abstracted summary which serves to identify the semantic structure of the object to which applies.
In P2T, protoforms include fuzzy temporal references for providing information about the temporal dimension of activities, arcs, or traces.
For example, the textual description \textit{Most executions of activity $\alpha_{1}$ last ten minutes in average more than those of activity $\alpha_{2}$} is generated by the protoform $Q \ B \ activity \ lasts \ A$. This is an activity-related protoform where $Q$ is the quantifier \textit{Most} $B$ is a qualifier, in this case, activity $\alpha_{1}$, and $A$ is the summarizer used to describe activity $\alpha_{1}$ \textit{ten minutes in average more than those of activity $\alpha_{2}$}.
Note that behavioral patterns can be expressed through relations between activities, meaning that pattern-related protoforms are compositions of arcs-related protoforms.
Protoforms have abstraction levels, as summarization does, this allows for a general abstracted summary to produce multiple different summaries depending on the knowledge used for its realization.

\subsubsection{Document planning and Realization}
Its objective is to generate the natural language descriptions of the process, taking the protoforms, templates, and expert knowledge as inputs.
In our model we follow a hybrid template-based realization, which uses some domain expert knowledge and is more rich and flexible than basic fill-in-the-gap approaches, but simpler and quicker than full fledged NLG system implementations. 
The SimpleNLG-ES \cite{aramossoto2017adapting} (Spanish version of the SimpleNLG realization engine) realization engine is used in this stage.

\section{Case study}
\label{sec:case-study}

We have applied our framework in a real case study in the health domain: the process related to the patients' management in the Valvulopathy Unit of the Cardiology Department of the University Hospital of Santiago de Compostela.
In this Unit, consultations and medical examinations, such as radiography, echocardiogram or Computed Tomography (CT) scans, are performed to patients with aortic stenosis in order to decide the treatment (including surgery) they will undergo.
Other information like unexpected events (e.g. emergencies, non-programmed admissions) and patient management activities (inclusion in the process, revisions, etc) are also recorded in the event log data.

Medical professionals have a real interest in applying process mining techniques to this process, since it allows them to extract valuable knowledge about the Unit like, relations between age, sex, admittance (emergency or normal admission) and the number of successful surgeries or delay between activities, the delays between crucial activities (such as the admission of a patient and its surgery) due to tests like CT scans or echocardiograms, the different paths of the process which patients with different attributes follow, etc.
The main goal to reach with all this information is to reduce the delays between process activities, prevent the repetition of activities (loops in the process), minimize patient management time, optimize resources and most important, increase the number of successful treatments.

\begin{figure}[!t]
 \centering
 \includegraphics[width=\linewidth]{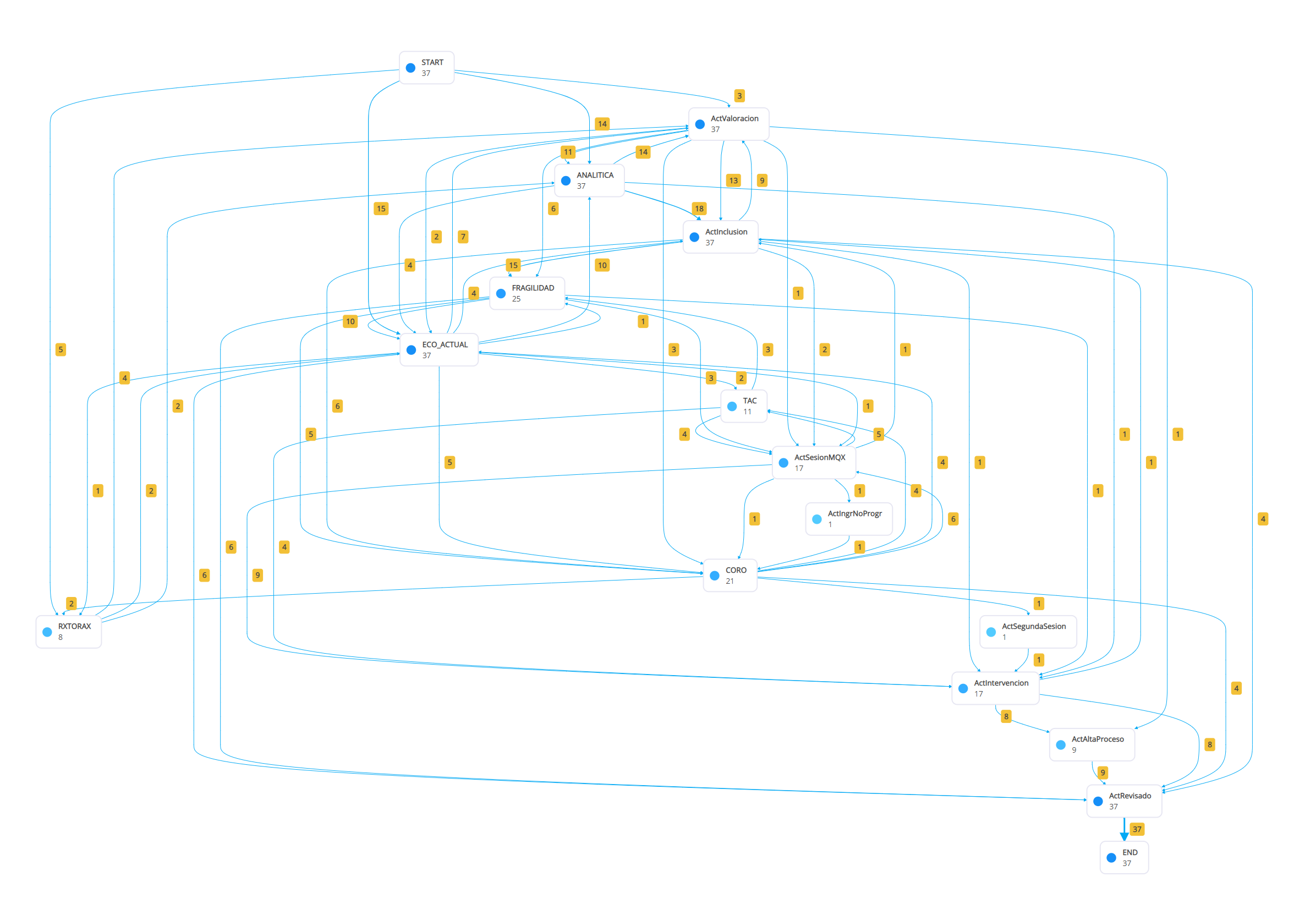}
 \caption{Model of the Valvulopathy process represented with the InVerbis Analytics visualization tools \cite{InVerbisAnalytics}.}
 \label{fig:process-model}
\end{figure}

In Fig. \ref{fig:process-model} the model that describes this process is shown, it has been discovered from an event log with data on 639 patients. 
 
Since medical, management activities and exceptions are recorded, and since patients' management depends on their pathological state, the frequency of each path (sequence of activities on a trace) in the process is very low (only the twenty most frequent paths from the six-hundred twenty-two total are shown in the figure) giving place to a highly complex model.
This makes it difficult for medical professionals to understand what happens within the process even when it is graphically represented.
To solve this, linguistic descriptions of the main process analytics are generated, as shown in Table  \ref{tab:descriptions}, facilitating the understanding of temporal relations and delays between activities and traces, which is the main concern of domain experts in this regard.

\begin{table*}[!t]
 \caption{Textual descriptions generated for the valvulopathy process}
 \begin{tabular}{p{12cm}}
  \hline
  
  - During the first half of year 2018, 52\% less Surgical Interventions were registered compared to the second half of that same year. period.                                                                                 \\
  
  - In the process, 78\% less Surgical Interventions than Coronographies were registered.                                                                                                                                       \\
  
  - Waiting time between Consultations is around 5 weeks and 6 days in average.                                                                                                                                                 \\
  
  - Waiting time between a Coronography and a CAT is around 6 weeks in average.                                                                                                                                      \\
  
  - Around 7 weeks and 3 days after the Medical-Surgical Session a patient undergoes Surgical Intervention.                                                                                                                     \\
  
  - 6\% of times, after the First  Medical-Surgical Session, a Second Session is held around 5 weeks and 3 days later. On the contrary, 33\% of times, patient undergoes Surgical Intervention around 7 weeks and 3 days later. \\
  
  - Patients who go through Assessment, Medical-Surgical Session and Surgical Intervention stay for 114 days in average at the cardiology service.                                                                              \\
  \hline
 \end{tabular}
 \label{tab:descriptions}
\end{table*}

\section{Conclusions and Future Work}
\label{sec:conclusions}

Our P2T framework integrates the process mining, natural language generation, and fuzzy linguistic protoforms paradigms for the automatic generation of textual descriptive explanations of processes, which include quantitative information (i.e., frequent and infrequent behavior, temporal distances between events and frequency of the relationships between events). A real use-case is presented, showing the potential of P2T for providing natural language explanations addressed to cardiology specialists about consultations and interventions of the patients of the valvulopathy unit.
As future work, extensive human validation of the generated descriptions will be conducted with domain experts. 

\section*{Acknowledgments}
\label{sec:acks}

This research was funded by the Spanish Ministry for Science, Innovation and Universities, the Galician Ministry of Education, University and Professional Training and the ERDF/FEDER program  (grants TIN2017-84796-C2-1-R, \linebreak ED431C2018/29 and ED431G2019/04).

\bibliographystyle{splncs04}
\bibliography{ref}

\end{document}